# Handwritten Chinese Character Recognition by Convolutional Neural Network and Similarity Ranking

Ludi Wang (M.Eng)
*Department of Electrical & Computer Engineer*
*University of Western Ontario*
lwang823@uwo.ca
250967311

Jinliang Zhang (M.Eng)
*Department of Electrical & Computer Engineer*
*University of Western Ontario*
jzhan964@uwo.ca
250919668

Junyi Zou (M.Eng)
*Department of Electrical & Computer Engineer*
*University of Western Ontario*
jzou44@uwo.ca
250833154

*Abstract* - Convolution Neural Networks (CNN) have recently achieved state-of-the art performance on handwritten Chinese character recognition (HCCR). However, most of CNN models employ the SoftMax activation function and minimize cross entropy loss, which may cause loss of inter-class information. To cope with this problem, we propose to combine cross entropy with similarity ranking function and use it as loss function. The experiments results show that the combination loss functions produce higher accuracy in HCCR. This report briefly reviews cross entropy loss function, a typical similarity ranking function: Euclidean distance, and also propose a new similarity ranking function: Average variance similarity. Experiments are done to compare the performances of a CNN model with three different loss functions. In the end, SoftMax cross entropy with Average variance similarity produce the highest accuracy on handwritten Chinese characters recognition.

*Key words* - Convolution Neural Networks (CNN); handwritten Chinese character recognition (HCCR); cross-entropy function; similarity ranking function.

## I. INTRODUCTION

This report is focused on the offline Handwritten Chinese character recognition, which is an important research field in pattern recognition. Recognizing Chinese characters is more challenging compared with alphabet and digits recognition. Because Chinese has over 7000 classes in the common vocabulary, and it also has more complex structure in each character. In recent years, Chinese handwriting character recognition has received much research and attention. As regular Neural network don't scale well to full image, in order to deal with the handwritten images, the most suitable approach is to use convolution neural network (CNN). The model in this report is based on CNN. However, unlike most existing methods, the method presented in this report using both classification and similarity ranking signals as supervision to train a CNN model. Classification is to classify an input image into a large number of identity classes, while similarity ranking is to minimize the intra-class distance while maximizing the inter-class distance.

The rest of this report includes the briefly reviews for convolutional neural network and cross-entropy, introduction of similarity ranking function, methodology of the experiment and the results analysis.

## II. BACKGROUND

### 2.1 Convolutional Neural Network

Convolutional neural networks (CNNs) consist of an input and an output layer, as well as multiple hidden layers. The hidden layers of a CNN typically consist of convolutional layers, pooling layers, fully connected layers and normalization layers [1]. Avoiding complex pre-processing of the image (extracting artificial features, etc.), CNNs process input original image directly. The layers of a CNN have neurons arranged in 3 dimensions: width, height and depth, CNN can reduce the full image into a single vector of class scores, arranged along the depth dimension through complete process [2]. In visual object recognition, CNNs often achieve a good performance. The convolution layer only increases the depth and leave the height and width unchanged. For the pooling layer, it would keep the depth, and narrow the size of volume. Finally, the fully connected layer would transform the volume into 1x1xn which n is represented to different classes [3]. Unlike the regular neural networks, CNN preserves the input's neighborhood relations and spatial locality in their latent higher-level feature representations. While the architecture of common fully connected neural network do not scale well to realistic-sized high-dimensional images in terms of computational complexity, CNNs do, since the number of free parameters describing their shared weights does not depend on the input dimensionality [4].

### 2.2 Loss Function for Classification Problems
### 2.2.1 Entropy

Entropy is a measure for information contents and could be defined as the unpredictability of an event. Assume that the probability of an event occurring is p, then the unpredictability of this event is $log_2(\frac{1}{p})log_2(\frac{1}{p})$ [5]. So, the greater the probability is, the smaller the unpredictability is, which means the information contents is also very small. If an event occurs inevitably with the probability of 100%, then the unpredictability and information content are 0.

Assume that a non-uniform dice, the probability of rolling it to get a definite point X ($X = \{x_1, x_2, ..., x_6\}$) is $p_i = p(X = x_i)$. The expectation of the discrete random variable can be defined as [5]:

$$H = -\sum_{i=1}^{n} p(x_i) \log p(x_i) = \sum_{i=1}^{6} p_i * \log(1/p_i)$$

The expectation H is entropy. So, the greater the entropy is, the greater the unpredictability is.

### 2.2.2 Cross-entropy

For same event, p is the real probability density, while q is the probability density from a prediction model. Cross-entropy can be defined as [6]:

$$H(p, q) = E_p\left[\frac{1}{\log(q)}\right] = -\sum_x p(x) \log[q(x)] = H(p) + D_{KL}(p||q)$$

$D_{KL}(p||q)$ is Kullback-Leibler Divergence. When $H(p)$ has a fixed value, $H(p, q)$ is equivalent to $D_{KL}(p||q)$,

$$D_{KL}(p||q) = \sum_i \left[p \log\left(\frac{p}{q}\right)\right]$$

which can be used to measure the similarity between p (true values) and q (values from predictor). When p=q, cross-entropy gives the minimum value, which is zero.

### 2.2.3 Cross-entropy Used as Loss Function

In machine learning and deep learning, cross-entropy can be used to define loss function. It represents the inaccuracy of predictions.

Considering a binary classification problem. There are only two classifications for the result of one prediction, so,

$$\begin{cases} p = \{y, 1-y\} \\ q = \{\dot{y}, 1-\dot{y}\} \end{cases}$$

cross-entropy is:

$$H(p, q) = E_p\left[\frac{1}{\log(q)}\right] = -\sum_x p(x) \log[q(x)] = -y \log(\dot{y}) - (1-y) \log(1-\dot{y})$$

which is the loss function of logistic regression. It computes the average cross-entropy of m samples [7]:

$$H(p, q) = -\frac{1}{m} \sum_{i=1}^{m} \sum_{j=0}^{k} H(p_j, q_j) = -\frac{1}{m} \sum_i [y_i \log(\dot{y}_i) + (1-y_i) \log(1-\dot{y}_i)]$$

$\dot{y}_i = p(y_i = j|x_i; \theta) = h(x_i; \theta)$ represents the probability of a prediction. The output of cross-entropy loss function is always greater than one. When the prediction is closer to the true value, the output of loss function is closer to zero.

Cross-entropy can be used in multiclassification problems with the combination of SoftMax (do not consider regularization):

$$J(\theta) = -\frac{1}{m} \sum_{i=1}^{m} \sum_{j=1}^{k} 1\{y_i = j\} \log \frac{e^{\theta_j^T x_i}}{\sum_{l=1}^{k} e^{\theta_j^T x_i}}$$

Compared with quadratic loss function, cross-entropy loss function gives better training performance on neural networks.

### 2.3 Similarity Ranking Function

In classification problems, besides differences in different classes, another important information can be learned is the similarities among different samples. Different similarity measurements may have impacts on the accuracy of the prediction.

### 2.3.1 Euclidean Distance

Euclidean Distance defines the true distance between two points in an m-dimensional space. In supervised learning, it can be used to measure the similarity between two samples with n features [8]:

$$d_{12} = \sqrt{\sum_{j=1}^{n} (x_{1j} - x_{2j})^2}$$

where $d_{12}$ refers to the similarity between first and second samples.

### 2.3.2 Average Variance Similarity

In the statistical description, variance is used to calculate the difference between each sample and the mean of all samples. It is defined as:

$$\sigma^2 = \frac{\sum_{i=1}^{m}(x_i - \mu)^2}{m}$$

where $\sigma^2$ is variance, $x_i$ is the i[th] sample, $\mu$ is mean of all samples, m is the number of samples.

### III. METHODOLOGY

The data used in this experiment is from CASIA online and offline Chinese handwriting databases. We selected data from the offline database. Original CASIA offline dataset is a list of images labeled with corresponding Chinese character, with size 6.7GB, contains 3160 different characters and total 1055440 samples, each character has average 334 corresponding samples. Due to the limitation of our machine, we only use a subset of the dataset, with 300 different characters and total 100342 samples. CASIA also provide offline character test dataset for evaluation, with average 81 different samples corresponding to each character. There are some handwriting samples from different writers shown in figure 1.

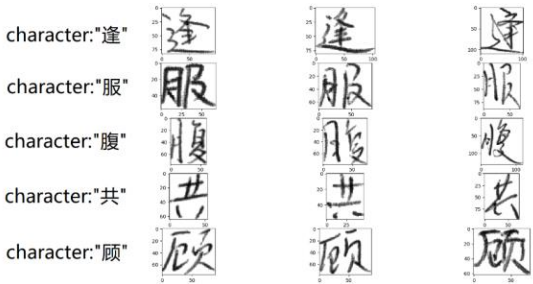

Fig. 1 Handwriting samples from 3 writers

*3.1 Data Preprocessing*

The original CASIA dataset is in "gnt" format, which is hard for TensorFlow to process, so we transform the dataset to "sqlite" format, which is easy to access through python package "sqlite3". Because the size of each image sample is different, so we perform cropping, scaling, and padding for each image, the result images have same size (128*128 pixels). Our group also perform normalization to the data, so the network is easier to converge in training period. There are some images for each character after preprocess in figure 2.

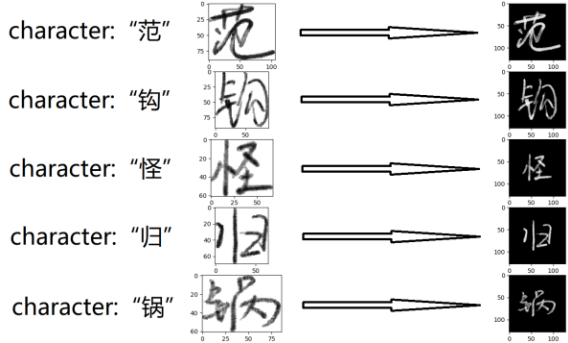

Fig. 2 Each character image after preprocess

*3.2 Model design*

Our group designed 3 different models, all 3 models have the same neural network structure, consists of 6 convolutional layers, 6 pooling layers, 2 fully connected layers and 2 dropout layers. For each convolutional layer and fully connected layer, we use leaky Relu as activation function to speed up the training process. However, they use different loss functions respectively. Model architecture can be summarized as the table 1.

*3.3 Validation process*

Due to the limitation of our machine, we didn't perform cross validation in our experiment, all training batch is randomly selected from the training dataset. However, there is another separated validation dataset provided by CASIA, which we only use in the validation process.

For model 1(SoftMax only), we use SGD (stochastic gradient descent) with batch size 200, every sample is random selected from the whole dataset. We use SoftMax Cross Entropy as the loss function to train model 1. SoftMax Cross Entropy can maximize the distance between different classes.

TABLE I
THREE MODELS OF CNNS WITH DIFFERENT LOSS FUNCTIONS

| MODEL A | MODEL B | MODEL C |
|---|---|---|
| Conv-32 | | |
| Pool | | |
| Conv-64 | | |
| Pool | | |
| Conv-128 | | |
| Pool | | |
| Conv-256 | | |
| Pool | | |
| Conv-256 | | |
| Pool | | |
| Conv-512 | | |
| Pool | | |
| FC1024 | | |
| Dropout 0.25 | | |
| FC1024 | | |
| Dropout 0.25 | | |
| FC300 | | |
| SoftMax | SoftMax plus Euclidean | SoftMax plus Variance |

For model 2(SoftMax plus Euclidean), we use batch size 180, with 90 characters and 2 samples for each character. Later use 2 samples of same class to calculate the Euclidean distance between them, we use Euclidean distance plus SoftMax Cross Entropy loss as the final loss function. Minimize the final loss function not only can minimize the Euclidean distance between same class, but also capable of minimize the SoftMax Cross Entropy loss.

For model 3(SoftMax plus variance), we use batch size 200, with 5 characters and 40 samples for each character. Later use 40 samples of same class to calculate the variance for each class, we use variance plus SoftMax Cross Entropy loss as the final loss function. Minimize the final loss function not only can minimize the variance between same class, but also capable of minimize the SoftMax Cross Entropy loss.

IV. EVALUATION AND RESULTS

This section first discusses the implementation details, then presents evaluation results comparing the proposed algorithm to 2 other competing models.

*4.1 Implementation details*

Our group use python package "pickle" and "sqlite" to manage dataset, which can access the database by database SQL. We implement all 3 different models with machine learning framework "TensorFlow", which is the new standard in deep learning industry. We choose "TensorFlow" because its capable of GPU acceleration. All 3 models are trained on GTX 1060 discrete GPU w/6GB GDDR5 graphics memory. It took 3 hours for model 1 to converge, and 5 hours for model 2 and model 3 to converge.

*4.2 Results*

After the training process, we perform evaluation process using another separated evaluation dataset. the recognition rates are applied to evaluate the accuracy of models with different loss functions. The results are shown in the table 2.

TABLE II
RECOGNITION RATES (%) ON DB1.1

| Loss function | Recognition rates |
|---|---|
| SoftMax | 93.79% |
| SoftMax + similarity ranking (Euclidean) | 94.99% |
| SoftMax + similarity ranking (Variance) | 95.58% |

For model without the usage of similarity ranking function, the recognition rate is 93.70%. After using Euclidean as similarity ranking function, the recognition rate is up to 94.99%. After using variance as similarity ranking function, the recognition rate is up to 95.58%.

Table 2 shows that the model with similarity ranking function has a small but consistent advantage compare to model without similarity ranking function. In addition, variance similarity ranking function is better than Euclidean similarity ranking function.

## V. CONCLUSION

Overall, handwritten Chinese character recognition is divided into two categories. One is online handwritten Chinese character, and the other is offline handwritten Chinese character. In this project, experiments are implemented on the dataset of offline handwritten Chinese character. This report focuses on the loss functions of a CNN model and demonstrates that the character classification and similarity ranking supervisory signals are complementary for each other, which can increase inter-class variations and reduce intra-class variations, and therefore achieve a much better classification performance. Combination of the two supervisory signals leads to significantly better results than only SoftMax cross-entropy based character classification. In addition, variance similarity ranking function have better performance than Euclidean similarity ranking function.